\documentclass[11pt,a4paper]{article}
\usepackage[hyperref]{acl2019}
\usepackage{times}
\usepackage{latexsym}
\usepackage{adjustbox}
\usepackage{graphicx}
\usepackage{subcaption}
\usepackage{comment}
\usepackage{algorithm}
\usepackage[noend]{algpseudocode}
\usepackage{amsmath}

\usepackage{xparse}
\NewDocumentCommand{\Log}{o}{%
  \IfNoValueTF{#1}{}{{}^{#1}\!}\log}%
\usepackage{amssymb}
\usepackage{framed}
\usepackage{url}
\usepackage{xcolor,colortbl}
\usepackage{booktabs}
\usepackage{mathtools, cases}
\usepackage{pifont}

\def\NoNumber#1{{\def\alglinenumber##1{}\State #1}\addtocounter{ALG@line}{-1}}

\usepackage{url}

\aclfinalcopy 

\title{BSDAR: Beam Search Decoding with Attention Reward in Neural Keyphrase Generation}

\author{Iftitahu Ni'mah\textsuperscript{\ding{168},\ding{171}} \hspace{1.5mm} Vlado Menkovski\textsuperscript{\ding{168}} \hspace{1.5mm} Mykola Pechenizkiy\textsuperscript{\ding{168}}    \\
{\normalsize{\textsuperscript{\ding{168}} Eindhoven University of Technology \hspace{1.5mm}\textsuperscript{\ding{171}} BRIN Indonesia}} \\
\texttt{\small{\{i.nimah, v.menkovski, m.pechenizkiy\}@tue.nl}}}

\date{}

\begin{document}
\maketitle
\begin{abstract}

This study mainly investigates two common decoding problems in neural keyphrase generation: sequence length bias and beam diversity. To tackle the problems, we introduce a beam search decoding strategy based on word-level and ngram-level reward function to constrain and refine Seq2Seq inference at test time. Results show that our simple proposal can overcome the algorithm bias to shorter and nearly identical sequences, resulting in a significant improvement of the decoding performance on generating keyphrases that are present and absent in source text. \footnote{The empirical study was completed in 2019 at Eindhoven University of Technology.}

 
\end{abstract}

\section{Introduction}

Neural keyphrase generation \cite{Meng2017Deep, Chen2018Keyphrase} studies a conditional Sequence-to-Sequence (Seq2Seq) learning \cite{sutskever2014sequence, cho2014learning, Jiatao2016Copying, See2017Get, vinyals2015show, xu2015show} in which a document source is paired to multiple keyphrases as its high level abstractions. Given document source as sequential inputs and the corresponding multiple keyphrase references as target sequential outputs, the training objective of neural keyphrase generation maximizes the averaged likelihood of correct keyphrase generation as next token prediction guided by attention mechanism.

Neural keyphrase generation also shares a similar objective with Seq2Seq-based neural summarization \cite{See2017Get} where the goal of the task is to condense a document into a short document abstraction. Consequently, both NLG domains also share common challenges. That is, the generation strategy and algorithm need to accommodate two mechanisms -- \textbf{to copy} words from source, and \textbf{to generate} semantically related words not featured in source document. While the copying or extracting task is particularly easy for an unsupervised keyword extractor (e.g. TfIdf), Seq2Seq mainly focuses on the generative objective and it has not been specifically trained on an extractive-typed task. This particular problem has been addressed by incorporating copying mechanism \cite{Jiatao2016Copying}, resulting in models referred as \textbf{CopyRNN} \cite{Meng2017Deep} and \textbf{CorrRNN} \cite{Chen2018Keyphrase}. However, although in general neural keyphrase generator outperforms traditional approaches when they are assessed based on retrieval evaluation metrics, the notably low decoding performance of Seq2Seq has not been further investigated.



%

\begin{figure*}[!t]
    \centering
    \includegraphics[scale=0.3]{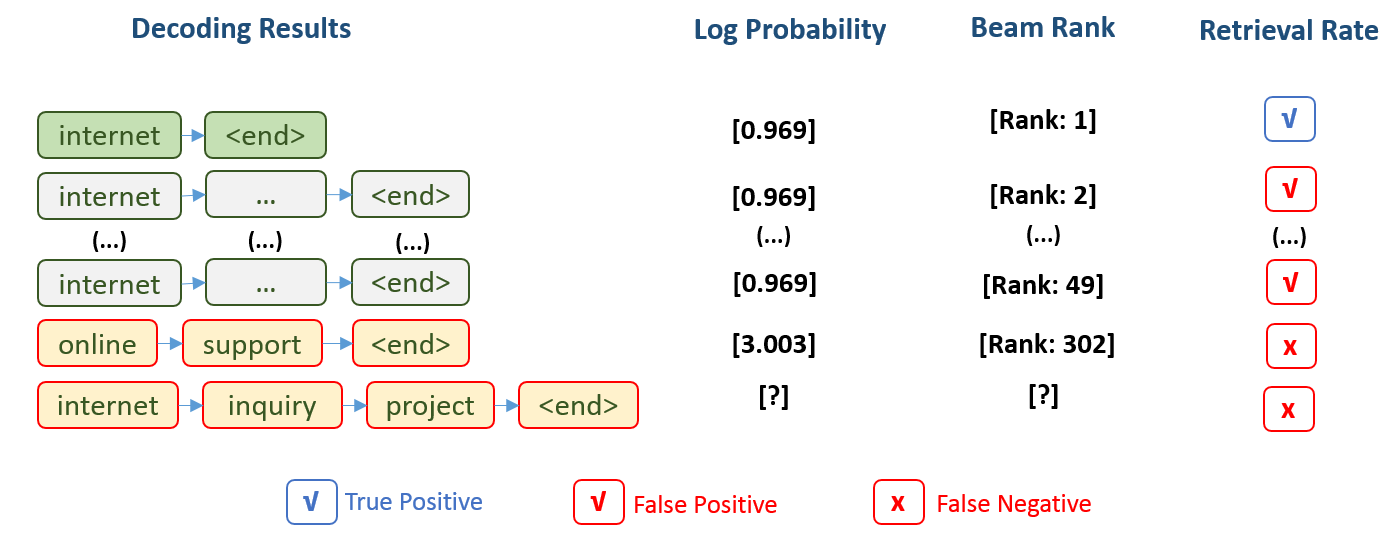}
    \caption{Text degeneration issues in neural keyphrase generation: sequence length bias and beam diversity.}
    \label{fig:bs_prob12}
\end{figure*}

\begin{figure*}[!ht]
    \centering
    \includegraphics[scale=0.3]{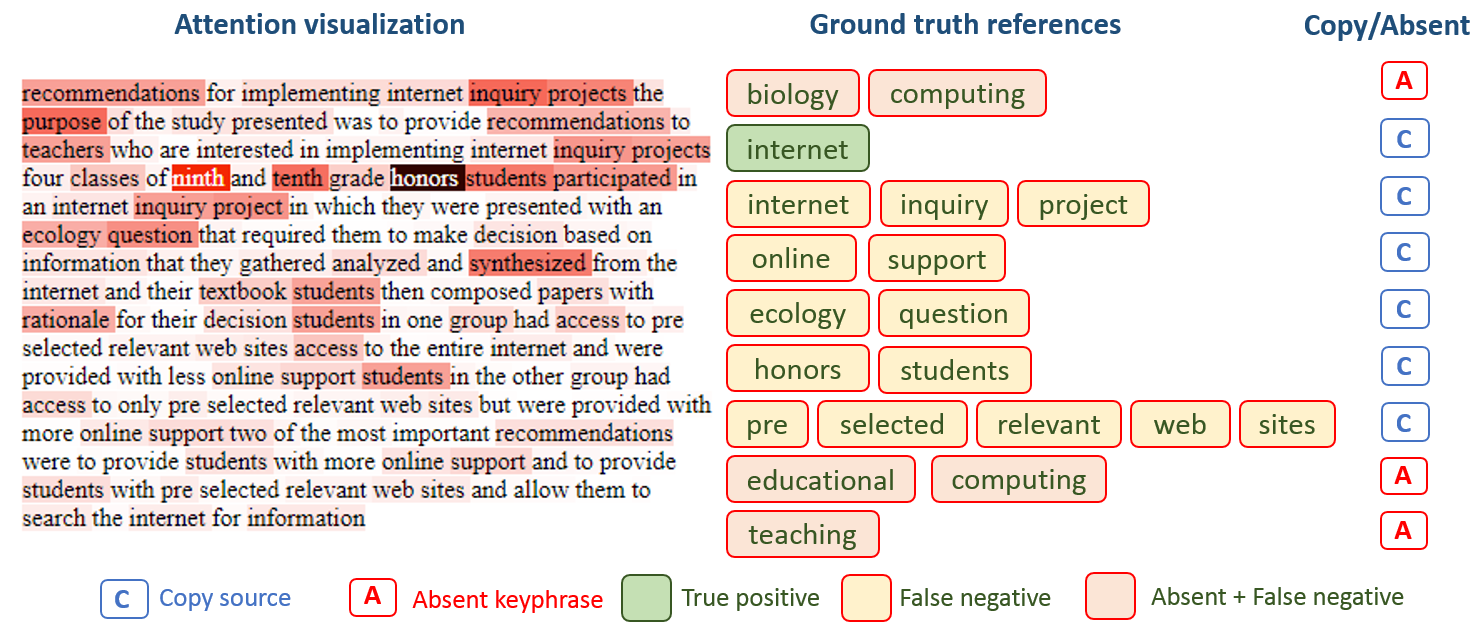}
    \caption{Verifying that the encoder network attended to influential words relevant to the ground truth keyphrase references.}
    \label{fig:copy_absent_kps}
\end{figure*}

Our work studies the decoding issues that are not specifically addressed and investigated in previous works on neural keyphrase generation. In particular, we focus on two common challenges of beam search decoding in diverse NLG tasks: (i) the algorithm bias to shorter sequences; and (ii) the algorithm tendency to produce nearly identical sequences, causing the beam candidates to be less diverse. Figure~\ref{fig:bs_prob12} illustrates an example of decoding results that have tendency to be shorter and repetitive. Whereas, Figure~\ref{fig:copy_absent_kps} shows a motivating example of performance discrepancy between what encoder and decoder part of Seq2Seq networks. In Figure~\ref{fig:copy_absent_kps} (left), the darker highlights correspond to higher attention scores given target keyphrases as query and word sequences in document source as keys. Although the encoder network has successfully attended on words relevant to the ground truth labels of keyphrases (right), the decoder network fails to generate the corresponding word sequences, as shown in Figure~\ref{fig:bs_prob12}. This example suggests that incorporating a good decoding strategy is as same important as ensuring the encoder network to learn a sufficiently good document and keyphrase representation as inputs for decoder network in inference stage.




\paragraph{Contribution}

To address the above text degeneration issues on sequence length bias and beam diversity, we introduce a mechanism to re-score the beam search decoding based on word-level and ngram level reward function conditioned on attention scores (section \ref{sec:bs_attention}). Our empirical results demonstrate that increasing the breath of beam size and the depth of sequences do not always provide optimal solutions in the current keyphrase generation task (section \ref{sec:empiric}).

\section{Related Work}
Previous works on Seq2Seq learning \cite{wiseman2016sequence, Ranzato2016Sequence} discuss two major text degeneration issues of Seq2Seq due to exposure bias and loss-mismatch between training and inference stage. To mitigate the issues, \citet{wiseman2016sequence} propose to incorporate beam search as training objective in addition to the standard negative log likelihood loss based on maximum likelihood estimation. Our study, however, focuses on the post-training solution. That is, by refining beam search decoding results through attention-based reward mechanism.

\section{Beam Search Decoding with Attention Reward (BSDAR)}
\label{sec:bs_attention}

\subsection{Word-level Attention-Reward}
We propose a word-level attention reward function \texttt{\small \textbf{WORD\_ATT\_REWARD}} as a mechanism to augment Seq2Seq prediction, shown in alg.\@ \ref{alg:word-att-rwd}. For each decoding time step $t$, the logits probability output of decoder network ($s_t$) is augmented by an attention vector ($\alpha_w \in R^{Tx}$), which each value corresponds to the attention weight of a word in source sequence $x$. $\hat{\alpha}_w[w]$ (alg. \ref{alg:word-att-rwd} \texttt{\small \textbf{line 3}}) denotes a normalized (mean) attention score of a particular word, given the possibility that the word occurs multiple times in the source text. Since the attention vector $\alpha_w$ is bounded to the sequential position of words in source input, the dictionary look-up of word index and the corresponding position $f(w): K \xrightarrow{} V$ is also given during the decoding time as a reference to calculate mean $\hat{\alpha}_w[w]$, where $K$ denotes a set of words in input sequence $K=\{w_0 \ldots w_N\}$ and $V$ corresponds to the position where the word occurs $V = [w_{p_1} \ldots w_{p_n}]$ in input sequence. Correspondingly, the augmentation of decoder logits only applies for words appear in source text. 

To intensify the effect of logits augmentation, which is becoming critical in a noisy decoding of Seq2Seq with large target vocabulary size (\textbf{CorrRNN-L}), we use $\lambda$ and $\gamma$ (alg. \ref{alg:word-att-rwd} line 4) as an augmentation factor of attention reward $\hat{\alpha}_w$. \emph{Here}, we set $\lambda = 2$ and $\gamma = \lambda * $ \texttt{\small \textbf{MAX}} $(\hat{\alpha}_w)$.

\begin{algorithm}[!ht]
  \caption{\textbf{\texttt{\small WORD\_ATT\_REWARD}}}
  \begin{algorithmic}[1]
    \For {$t=0, 1, \ldots, $ \texttt{\small MAX\_STEPS}}
    \State Collect logits and attention weights
    \NoNumber{$s_t = $ \textbf{\texttt{\small DEC}} $(y_{t}, c_{t-1}^{(\texttt{\small cov})}, c_{t-1}^{(\texttt{\small rev})}, s_{t-1}, \zeta_{t-1})$}
    \NoNumber{$\alpha = \textbf{\texttt{\small MLP}} (s_{t-1}, h, \texttt{\small cov})$}
    \NoNumber{$s_t \in R^{V}, \alpha \in R^{T_x}$}
    \State Compute normalized $\hat{\alpha}_w$, for each word 
    \NoNumber{$\hat{\alpha}_w[w] = \texttt{\small \textbf{MEAN}}(\alpha[w_{p_1} \ldots w_{p_n}])$}
    \State Augment the logits, for words in $\hat{\alpha}_w$
    \NoNumber{$\widetilde{s}_t[w] = s_t[w] + (\lambda * \hat{\alpha}_w[w]) + \gamma $}
    \State Get $b-$top most probable words 
    \NoNumber{\texttt{\small tokens} $=$ \textbf{\texttt{\small ARGSORT}} $\{\widetilde{s}_t\}[:b]$}
    \NoNumber{\texttt{\small probs} $=$ \textbf{\texttt{\small SORT}} $\{\widetilde{s}_t\}[:b]$}
    \State \textbf{return} \texttt{\small BEAM-HYP} $=$ \texttt{\small (tokens, probs)}
    \EndFor
  \end{algorithmic}
 \label{alg:word-att-rwd}
\end{algorithm}

\begin{algorithm*}[!t]
  \caption{\textbf{\texttt{\small NGRAM\_ATT\_REWARD}}}
  \begin{algorithmic}[1]
    \While{\texttt{\small steps} $<$ \texttt{\small MAX\_STEPS} \textbf{and} \texttt{\small Results} $<$ \texttt{\small BEAM\_SIZE} $b$}
    \State Run one step decoding with word attention reward
    \NoNumber{\texttt{\small BEAM-HYP} $=$\texttt{\small \textbf{WORD\_ATT\_REWARD}()}}
    \State Expand tree with new candidates \texttt{\small HYPS} $\frown$\texttt{\small BEAM-HYP}
    \State Construct sequence candidates
    \NoNumber{\texttt{\small SEQ} $=$ \texttt{\small HYPS.PREV\_TOKENS} $+$ \texttt{\small HYPS.CURR\_TOKENS}}
    \If{\texttt{\small SEQ} found in annotated source \texttt{\small ATT-ANNOT}}
    \State Reward candidates with $\hat{\alpha}_p$
    \NoNumber{\texttt{\small AUG\_PROB} $=$ \texttt{\small HYPS.CURR\_PROB}  $ + (\lambda * \hat{\alpha}_p) + \gamma $}
    \ElsIf{\texttt{\small SEQ} is partly composed of tokens in \texttt{\small ATT-ANNOT}}
        \State Set probability into negative value (log ``inf'') to penalize \texttt{\small SEQ} candidate
        \NoNumber{\texttt{\small AUG\_PROB} $= -0.05$}
    \Else
        \State Set logits probability without reward score 
        \NoNumber{\texttt{\small AUG\_PROB} $=$ \texttt{\small HYPS.CURR\_PROB} }
    \EndIf
    \State Re-sort beam candidates 
    \State Update tree with new candidates
    \State Sort the candidates stored in memory (tree) based on normalized joint probability
    \State Expand \texttt{\small Results} with completed sequence candidates 
    \EndWhile
  \end{algorithmic}
 \label{alg:ngram-att-rwd}
\end{algorithm*}

Our proposed word-level attention reward shares a common intuition with actor-critic algorithm for sequence prediction \cite{Bahdanau2017Actor}. Intuitively, the \texttt{\small \textbf{WORD\_ATT\_REWARD}} function corresponds to increasing the probability of \emph{actions} that the encoder network gives a higher attention value; and decreasing the probability of \emph{actions} that are being ignored by the encoder network. The actions correspond to the Seq2Seq predictions for each decoding step. The proposed reward mechanism, thus, bounds the actions with an attention score, which each value represents a degree of importance of word in source sequence based on what the Seq2Seq attention network has learnt during training time.

\subsection{$N-$gram-level attention reward} 

The proposed word-level attention reward in alg. \ref{alg:word-att-rwd} is based only on bag-of-words assumption. As a potential trade-offs on the decoding performance, the beam algorithm may then favour sequence candidates containing words with higher attention score. Thus, it disregards whether the formed sequence is sensible as a candidate of keyphrase. For instance, from the attention visualization in fig. \ref{fig:copy_absent_kps}, a keyphrase candidate ``\texttt{\small inquiry honours}'' may also be rewarded highly, considering the sequence is composed of words with a high attention score (``\texttt{\small inquiry}'', ``\texttt{\small honours}''). This leads to a noisy prediction and can potentially decrease decoding performance.

To mitigate this issue, we also introduce the $n-$gram-level attention reward function \texttt{\small \textbf{NGRAM\_ATT\_REWARD}} to further augment and penalize beam candidates before adding them into the memory (alg. \ref{alg:ngram-att-rwd}). For each decoding step $t$, given the previous stored tokens in memory and the current beam candidate returned by \texttt{\small \textbf{WORD\_ATT\_REWARD}}, an $n-$gram candidate \texttt{\small SEQ} were formed. For all \texttt{\small SEQ} candidates matched against the extracted $n-$gram attention annotations \texttt{\small ATT-ANNOT}, the logits of the last tokens of the \texttt{\small SEQ} candidates were added by the corresponding $n-$gram attention score $\hat{\alpha}_p$ (alg. \ref{alg:ngram-att-rwd} line 6).

\paragraph{Automated attention-based annotation}
The steps for acquiring both extracted n-gram attention annotations \texttt{\small ATT-ANNOT} and the corresponding attention score are shown in alg. \ref{alg:att-annot}. The attention annotation \texttt{\small ATT-ANNOT} was constructed during the first decoding step ($t=0$) by a simple $n-$gram ($n=1, \ldots, 5$) chunking method (i.e. sliding window of $n$ and $n+1$) of a filtered source document. The filtering of source document is based on a percentile rank $10$ of sorted attention score $\alpha$ as a cutting attention threshold $\tau$. Given the attention threshold $\tau$, the \texttt{\small ATT-ANNOT} was extracted based on $n-$gram chunks of element-wise multiplication between filtered attention score $\alpha^{\delta}$ (\texttt{\small \textbf{line 3}}) and source sequence (\texttt{\small \textbf{line 4}}). The final result is a list of $n-$grams with the corresponding attention score $\hat{\alpha}_p$. $\hat{\alpha}_p$ was acquired by computing the mean of attention scores of words $p_i$ composing the corresponding $n-$gram sequence $P$:  $\hat{\alpha}_p[P] = \frac{1}{N_p} \sum_{i=1}^{N_p} \hat{\alpha}[p_1] + \ldots + \hat{\alpha}[p_i] $. The resulting extracted atention annotation from the first decoding step is then used as a global reference for the following decoding time steps ($t>0$) and is utilized as reference for \texttt{\small \textbf{NGRAM\_ATT\_REWARD}} function.

\begin{algorithm}[!ht]
  \caption{\textbf{\texttt{\small EXTRACT\_ATT\_ANNOT}}}
  \begin{algorithmic}[1]
    \State Collect attention vector $\alpha$
    \State Compute threshold $\tau$ based on percentile $10$ of $\alpha$
    \State Binarize attention values based on threshold $\tau$ 
    \NoNumber{$\alpha^{\delta}_{t} = [1$ if $\alpha_t>\tau$ else $-1]$}
    \State Extract n-grams attention annotation 
    \NoNumber{\texttt{\small ATT-ANNOT} = \texttt{\small \textbf{CHUNK}}$(\alpha^{\delta} * x_{1 \ldots T_x})$}
    \State Extract $\alpha$ based on sequential position
    \NoNumber{\texttt{\small SEQ-POS} = \texttt{\small \textbf{CHUNK}} $(\alpha^\delta *$ $x-$\texttt{\small pos}$)$}
    \NoNumber{$\hat{\alpha}[P] = \alpha[$\texttt{\small SEQ-POS}$_i]$}
    \State Extract n-gram attention values $\hat{\alpha}_p$ 
    \NoNumber{$\hat{\alpha}_p =$ \texttt{\small \textbf{MEAN}} $(\hat{\alpha}[P_1 \ldots P_n])$}
    \State \textbf{Return} \{\texttt{\small ATT-ANNOT} $= f: K \xrightarrow{} V $, $K=\{P_1 \ldots P_n\}$ $V = [ \hat{\alpha_p} ]$\}
  \end{algorithmic}
 \label{alg:att-annot}
\end{algorithm}

\paragraph{Penalty score}
Penalty was given to the beam candidates that are partially composed of word tokens found in \texttt{\small ATT-ANNOT}. These sequence candidates contain words with high attention score, but are mainly \emph{non-}sensical. In this subset, the last tokens of the sequence candidates were set with a negative probability ($-0.05$), shown in alg. \ref{alg:ngram-att-rwd} \texttt{\small \textbf{line 8}}. For all candidates with negative value of probability in beam tree, the logits were then set to zero. This penalty encourages the sequence candidates to have an extremely large log probability values (``inf''), and correspondingly lower ranks during beam re-sorting stage. For sequence candidates that do not contain words and phrases in \texttt{\small ATT-ANNOT}, the logits output of decoder network were not augmented, kept \emph{as is} (\texttt{\small \textbf{line 10}}). Thus, the sequences not featured in source text were still considered as candidates of Seq2Seq final prediction, but ranked after those found in the source sequence. This last step intuitively aims to preserve the ``abstractive'' ability of the model, i.e. the ability to produce sequences that do not necessarily appear in the source but has a close semantic meaning with the corresponding document.

\subsection{Re-rank method}
\label{sec:heuristic2}

In addition to the proposed decoding with attention reward, a heuristic approach to alleviate sequence length bias and diversity issues was employed. We adopted a concept of \emph{intra-} and \emph{inter-} sibling rank of beam decoding \cite{LiJurafskiMutual2016} into a simple implementation. We refer the heuristic approach adopted in this study as (1) \emph{pre-}intra siblings rank; (2) \emph{post-}intra siblings re-rank; and (3) \emph{post-}inter sibling re-rank. \emph{Here}, in \emph{pre-}intra siblings ranking, for each decoding step, we only consider top-3 beams (word tokens) to be added into the tree queue. Given completed sequences (i.e. sequences with ``\texttt{\small <end>}'' as last token), in \emph{post-}intra siblings re-rank, given candidates with the same parent node and sequence length, only top-1 beam candidates were considered. Likewise, in \emph{post-}inter sibling re-rank, only top-5 candidates were considered as final solution. While \emph{pre-}intra siblings rank was ranked based on the probability scores of the last tokens of the sequence candidates, the \emph{post-}intra siblings re-rank and \emph{post-}inter siblings re-rank were sorted based on normalized joint probability score of the completed sequences.


\vspace{-1em}
\section{Increasing traversal breadth and depth}
\label{sec:empiric}

Conventional wisdom for tackling the aforementioned beam search decoding issues is by expanding the beam size (traversal breadth) and sequence length (traversal depth). \emph{Here}, we show empirically that increasing both parameters does not guarantee to result optimal solutions or significantly increase the performance gain in the current study.

\paragraph{Expanding Beam Size}

\begin{figure}
    \centering
    \includegraphics[width=.78\linewidth]{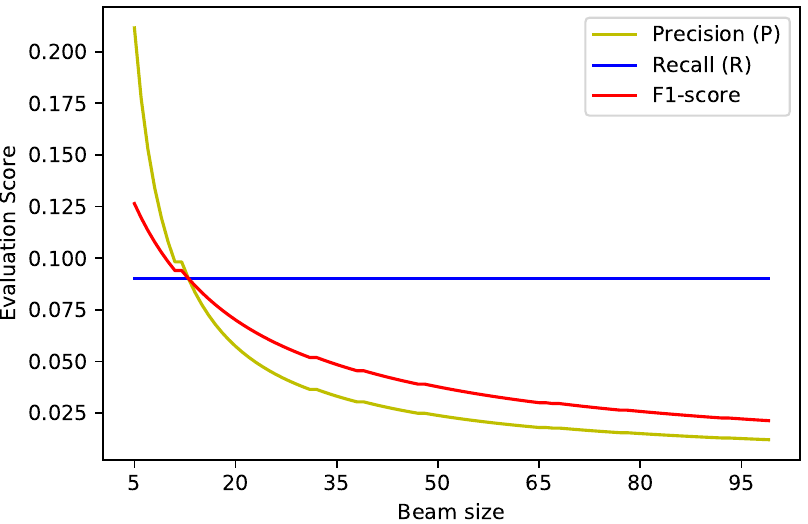}
    \caption{$n-$gram evaluation}
    \label{fig:bs1_a}
    \vspace{-1em}
\end{figure}

\begin{figure}
    \centering
    \includegraphics[width=.78\linewidth]{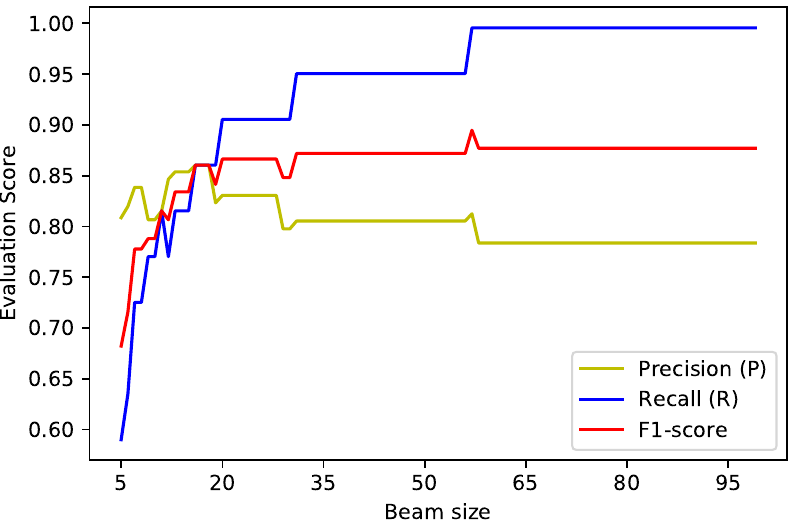}
    \caption{uni-gram evaluation}
    \label{fig:bs1_b}
    \vspace{-1em}
\end{figure}

Figure (\ref{fig:bs1_a} shows that there is no gain on increasing beam size (up to $95$ beams) and utilizing a simple length normalization technique. The different trend of uni-gram-based evaluation in figure \ref{fig:bs1_b}, however, indicates that the beam solution includes potential candidates partially (i.e. contains partial words in references), but fails to correctly form $n-$gram sequence candidate longer than a uni-gram. Example of the decoding results is shown in table \ref{table:gen_bs1}.



\begin{table}[!ht]
\centering
\resizebox{.45\textwidth}{!}{
\begin{tabular}{ | c | c | }
\hline 
\bf No Length Normalization & \bf With Length Normalization \\
 \hline
 internet & internet\\
 recommendations & internet analysis\\
 support & recommendations\\
 $<$unk$>$ & support\\
 online & $<$unk$>$ \\
 information & online\\
 internet analysis & information\\
 web & web\\
 decision & decision\\
 computer & computer\\
\hline
\end{tabular}
}
\caption{\label{table:gen_bs1} Decoding results. Beam size $b$ is set to 10.} 
\vspace{-1em}
\end{table}



\vspace{-1em}
\paragraph{Increasing sequence length}
\label{sec:seqlen}

\begin{figure}
    \centering
    \includegraphics[width=.78\linewidth]{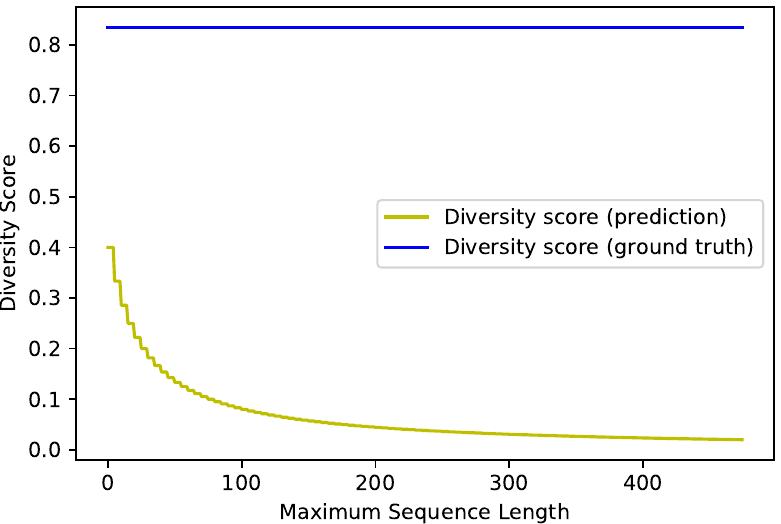}
    \caption{\texttt{\small MAX\_SEQ} vs. diversity score}
    \label{fig:bs1_d}
    \vspace{-1em}
\end{figure}

Increasing the traversal depth (\texttt{\small MAX\_SEQ}), shown in figure \ref{fig:bs1_d}, also does not add any values on improving the diversity of beam candidates in the current task. We define a diversity score (\emph{y} axis of figure \ref{fig:bs1_d}) as the ``diversity'' of the first word tokens in prediction and ground truth set, as such $\texttt{\small \textbf{DIV\_SCORE}} = \frac{\texttt{\small \textbf{COUNT}} (\{w^{(1)}\})}{\texttt{\small \textbf{COUNT}}(Y)}$. Where $\{w^{(1)}\}$ denotes a set of unique words, which each corresponds to the first token of a sequence and $Y$ corresponds to a list of keyphrases. The purpose of this diversity score metric is to measure the repetitiveness of beam decoding based on the first word tokens of the generation results. The higher the diversity score is (e.g. close to the diversity score of ground truth references), the better of the decoding algorithm to overcome the beam diversity issue.

\section{Experiments}
\label{sec:bs_issues}



\subsection{Neural Keyphrase Generation}

\paragraph{Seq2Seq Attention Model}
Our Attention-based Seq2Seq model is constructed of: \textbf{(1)} \textbf{encoder - decoder} architecture with \textbf{attention mechanism} \cite{Bahdanau2014Neural} as a backbone architecture; \textbf{(2)} \textbf{copying mechanism} \cite{Jiatao2016Copying, See2017Get, Meng2017Deep}; and \textbf{(3)} \textbf{coverage-review mechanism} \cite{See2017Get, Chen2018Keyphrase}. We re-implemented and modified \textbf{CopyRNN} \cite{Meng2017Deep} and \textbf{CorrRNN} \cite{Chen2018Keyphrase} to be able to be trained on two conditions of target vocabulary size: truncated vocabulary and a very large target vocabulary \cite{jean2015Using}. In truncated vocabulary setting, the vocabulary included in look-up dictionary is constrained to top-50K most frequent words, while Out-of-Vocabulary (OOV) set is referred to as \texttt{\small {<unk>}}.  In this study, Seq2Seq model with truncated vocabulary is referred as \textbf{CorrRNN}, while the modified model for large target vocabulary is referred as \textbf{CorrRNN-L}.


\paragraph{CorrRNN}
For both Seq2Seq models in this study (CorrRNN, CorrRNN-L), we use MLP (concat) attention mechanism \cite{Bahdanau2014Neural} as the attention scoring function \texttt{\small score}$($\texttt{\small Q,K}$) = $\texttt{\small V}$^T$ \texttt{\small tanh}$($\texttt{\small W}$[$\texttt{\small Q,K}$])$ for incorporating copying, coverage, and review mechanism into Seq2Seq architecture. \texttt{\small Q} corresponds to query-attention: decoder state at one time step $s_t$; and \texttt{\small K} denotes keys-attention: a sequence of encoder states $h_{1 \ldots T_x}$ and coverage vector of attention \texttt{\small cov} up to the current time step $t$. The latter corresponds to a coverage mechanism \cite{See2017Get,Chen2018Keyphrase}. \emph{Here}, for \textbf{CorrRNN}, the model has two inputs: (1) sequences with a truncated dictionary size referred as $x$; and (2) sequences with extended vocabulary referred as $x^{(ext)}$. The additional \texttt{\small oov} dictionary index size is set to $23914$, excluding the main vocabulary index with top$-50K$ most frequent words. 

\paragraph{CorrRNN-L}
Since the vocabulary is not being truncated, CorrRNN-L model does not have issues with OOV. The problem, however, is shifted to the complexity of training and decoding due to large number of target words. This issue has been addressed by incorporating sampled softmax approximation based on adaptive sampling algorithm \cite{jean2015Using} as an approximate learning approach to a very large target vocabulary size. \emph{Here}, we trained the model on $366730$ vocabulary size. The number of vocabulary samples for the approximate softmax layer, instead of \emph{full} softmax in \textbf{CorrRNN} model architecture, is set to $50K$.

\subsection{Beam Search Decoding Issues}

\paragraph{Sequence Length Bias}

By default, beam search decoding algorithm stops the search when exactly $b$ completed candidates of target sequences found, i.e. when the decoder network generates {\small{\texttt{$<$end$>$}}} token. As illustrated in fig.\@ \ref{fig:bs_prob12}, although increasing beam size increases the possibility of more candidates to explore, the likelihood for the model to find the {\small{\texttt{$<$end$>$}}} token after the first decoding step is also high. Consequently, the beam mainly contains short sequences ($1-$grams, $2-$grams), disregarding many potential $n-$grams candidates with longer sequences ($n \geq 3$). This tendency of the decoding algorithm to favour shorter sequences can hurt performance severely, specifically in the current keyphrase generation task, where ground truth references are in variable length ($n-$grams, $n = 1, \ldots, 5$). We further show empirically in section \ref{sec:empiric} that solely utilizing a normalization technique does not guarantee to solve the sequence length bias issues in the current task.




\paragraph{Beam Diversity}

Figure \ref{fig:bs_prob12} exemplifies diversity problem in the standard beam search decoding of Seq2Seq model for the current task. The generation of nearly identical beams, i.e. $>80\%$ of sequences started with word {\small{\texttt{internet}}}, results in a low \emph{informativeness} (Precision $P$). Consequently, it decreases the decoding performance in the current task. 

\subsection{Hyperparameters}

Seq2Seq model in this study was trained on KP20k corpus \cite{Meng2017Deep} ($\approx 530K$ documents, providing $\approx 2M$ training sets after preprocessing). For training the model, both sources $x_i$ and the corresponding keyphrase labels $P_i^{(n)}$ are represented as word sequences, such as $x_i = \{x_1^{(i)}, x_2^{(i)}, \cdots x_{T_x}^{(i)}\}$ and $P_{i,n} = \{p_1^{(i,n)}, p_2^{(i,n)}, \cdots p_{T_y}^{(i,n)}\}$, where $T_x$ and $T_y$ are maximum sequence length of source document and target keyphrase labels respectively. To be noted, each source document corresponds to $n$ multiple keyphrase labels. By splitting the data sample $\{x_i,P_i\}$ into $n$ pairs $\{x_i,P_i^{(1)}\}, \ldots, \{x_i,P_i^{(m)}\}$, the training set is presented as text-keyphrase pairs, each contains only one source text sequence and target keyphrase sequence. In inference stage, standard evaluation data sets for keyphrase extraction were used (Inspec \cite{Hulth:2003} ($201$), Krapivin \cite{Krapivin:2009} ($201$), NUS \cite{Nguyen:2007} ($156$), Semeval-2010 \cite{Kim:2010} ($242$).

\paragraph{Beam Parameters}
For the first decoding time step ($t=0$), beam size ($b$) is set to be a larger number ($b=100$) than the remaining decoding time steps ($b=50$). Number of hypothesis \texttt{\small num\_hyps} representing the partial solutions (queue) of sequence candidates to be added into the memory is set to $200$. 

\paragraph{Evaluation Metrics}
We employ Micro-average Recall (R), ROUGE-L average F1-score, and diversity metrics to evaluate performance of beam search decoding algorithms in this study.

\section{Results and Discussion}
\label{sec:exp}



\begin{table*}[!ht]
\centering
\resizebox{.75\textwidth}{!}{
\begin{tabular}{ | l | cc cc cc cc |}
\hline
\bf BS Decoding & \multicolumn{8}{|c|}{\bf CorrRNN} \\
& \multicolumn{2}{c}{\bf Inspec (201)$^{\dagger}$} & \multicolumn{2}{c}{\bf Krapivin (201)$^{\dagger}$} & \multicolumn{2}{c}{\bf NUS (156)$^{\dagger}$} & \multicolumn{2}{c|}{\bf Semeval-2010 (242)$^{\dagger}$}  \\
& \bf $R@50$ & \bf $R@N$ & \bf $R@50$ & \bf $R@N$ & \bf $R@50$ & \bf $R@N$ & \bf $R@50$ & \bf $R@N$ \\
 \hline
 BS & 0.098 & 0.098 & 0.073 & 0.073 & 0.093 & 0.093 & 0.056 & 0.056  \\
 BS++ & 0.091 & 0.091 & 0.094 &  0.094 & 0.098 & 0.098 & 0.059 & 0.059  \\
\bf \textit{BSDAR} & \bf 0.290 & \bf 0.372  & \bf 0.249 & \bf 0.317  & \bf 0.211 & \bf  0.255 & \bf 0.169 & \bf  0.221\\
\hline
\end{tabular}
}
\caption{\label{table:bs_corrRNN} Comparison of Beam Search (BS) decoding algorithms on CorrRNN performance. \small{\textit{$^{\dagger}$ size of test corpora}}} 
\end{table*}

\begin{table*}[!ht]
\centering
\resizebox{.75\textwidth}{!}{
\begin{tabular}{ | l | cc cc cc cc |}
\hline
\bf BS Decoding & \multicolumn{8}{|c|}{\bf CorrRNN-L} \\
& \multicolumn{2}{c}{\bf Inspec (201)$^{\dagger}$} & \multicolumn{2}{c}{\bf Krapivin (201)$^{\dagger}$} & \multicolumn{2}{c}{\bf NUS (156)$^{\dagger}$} & \multicolumn{2}{c|}{\bf Semeval-2010 (242)$^{\dagger}$}  \\
& \bf $R@50$ & \bf $R$ & \bf $R@50$  & \bf $R$ & \bf $R@50$ &  \bf $R$ & \bf $R@50$ & \bf $R$ \\
 \hline
 BS & 0.045 & 0.045 & 0.038 &  0.038 & 0.043  & 0.043 & 0.026 & 0.026 \\
 BS++ & 0.033 &  0.047 & 0.038 & 0.05 & 0.0409 & 0.052 &  0.020 & 0.029 \\
\bf \textit{BSDAR} & \bf 0.268 &  \bf 0.327 & \bf 0.193 &  \bf 0.226 & \bf 0.152 &   \bf 0.182 & \bf 0.139 & \bf 0.166 \\
\hline
\end{tabular}
}
\caption{\label{table:bs_corrRNN_L} Comparison of Beam Search (BS) decoding algorithms on CorrRNN-L performance. \small{\textit{$^\dagger$ size of test corpora}}}
\end{table*}

\begin{table*}[!ht]
\centering
\resizebox{\textwidth}{!}{
\begin{tabular}{ | l | cccc | cccc | cccc | cccc |}
\hline
\bf Decoding & \multicolumn{4}{c|}{\bf Inspec (199)$^{\dagger}$} & \multicolumn{4}{c|}{\bf Krapivin (185)$^{\dagger}$} & \multicolumn{4}{c|}{\bf NUS (148)$^{\dagger}$} & \multicolumn{4}{c|}{\bf Semeval-2010 (242)$^{\dagger}$}  \\
& \multicolumn{2}{c}{\bf CorrRNN} & \multicolumn{2}{c|}{\bf CorrRNN-L} & \multicolumn{2}{c}{\bf CorrRNN} & \multicolumn{2}{c|}{\bf CorrRNN-L} & \multicolumn{2}{c}{\bf CorrRNN} & \multicolumn{2}{c|}{\bf CorrRNN-L} & \multicolumn{2}{c}{\bf CorrRNN} & \multicolumn{2}{c|}{\bf CorrRNN-L} \\
&\bf {\small R} & \bf {\small ROUGE-L} &\bf {\small R} & \bf {\small ROUGE-L} &\bf {\small R} & \bf {\small ROUGE-L} &\bf {\small R} & \bf {\small ROUGE-L} &\bf {\small R} & \bf {\small ROUGE-L} &\bf {\small R} & \bf {\small ROUGE-L} &\bf {\small R} & \bf {\small ROUGE-L} &\bf {\small R} & \bf {\small ROUGE-L} \\
 \hline
 BS & 0.027 & 0.207 & 0.066 & 0.211 & 0.026 & 0.257 & 0.053 & 0.226 & 0.034 & 0.240 & 0.048  & 0.197 & 0.015  & 0.194 & 0.026 & 0.154\\
 BS++ & 0.022 & 0.194 &  0.078 & 0.329 & 0.024 & 0.247 & \bf 0.081 & \bf 0.369 & 0.029 & 0.228 & 0.064  & 0.304 & 0.013 & 0.182 & 0.031 & \bf 0.260\\
\bf \textit{BSDAR} & \bf 0.038 & \bf 0.249 & \bf 0.079  & \bf 0.331 & \bf 0.071 & \bf 0.300 & 0.064  & 0.348 & \bf 0.037 & \bf 0.260 & \bf 0.065   & \bf 0.310 & \bf 0.031 & \bf 0.225 & \bf 0.041 & 0.253 \\
\hline
\end{tabular}
}
\caption{\label{table:ABS1} Abstractive Performance. Scores are based on \texttt{\small{Micro-Average Recall (R)}} and \texttt{\small ROUGE-L average F1-score}. \small{\textit{$^{\dagger)}$ size of test corpora after preprocessing.}}}
\end{table*}

\subsection{Decoding performance}

Table \ref{table:bs_corrRNN} and \ref{table:bs_corrRNN_L} shows the comparison between our proposed method (\textit{\textbf{BSDAR}}) and standard beam search decoding algorithms for Seq2Seq prediction. To show that simply applying the heuristic rules (section \ref{sec:heuristic2}) does not guarantee to solve the decoding issues, we included the heuristic-based only beam decoding (BS++) as a comparison. In general, the results for both \textbf{CorrRNN} and \textbf{CorrRNN-L} models show that \textit{\textbf{BSDAR}} can significantly improve the recall score of about $20-30 \%$. 

We also show that while expanding retrieval set (from $R@50$ to $R@N$, where $N=50 \ldots 200$) has no gain for the decoding based on standard beam decoding (BS) and heuristic-based decoding (BS++), the performance of the proposed decoding method significantly increases. This result indicates that the solutions based on standard beam search (BS) and heuristic-based beam search (BS++) mainly contain noises and non-relevant sequences, as compared to the predictions resulted by \textit{\textbf{BSDAR}}.

\begin{table*}[!ht]
\centering
\resizebox{\textwidth}{!}{
\begin{tabular}{ | l | cccc cccc |}
\hline
\bf Decoding & \multicolumn{2}{c}{\bf Inspec (138)$^{\dagger}$} & \multicolumn{2}{c}{\bf Krapivin (85)$^{\dagger}$} & \multicolumn{2}{c}{\bf NUS (93)$^{\dagger}$} & \multicolumn{2}{c|}{\bf Semeval-2010 (212)$^{\dagger}$}  \\
& \bf CorrRNN & \bf CorrRNN-L & \bf CorrRNN & \bf CorrRNN-L & \bf CorrRNN & \bf CorrRNN-L & \bf CorrRNN & \bf CorrRNN-L \\
 \hline
 BS & 0.236  & 0.189 & 0.236 & 0.183  & 0.215  & 0.150 & 0.196  & 0.142 \\
 BS++ & 0.249  & 0.328 & 0.248  & 0.344  & 0.234 & 0.279 & 0.199  & 0.269 \\
\bf \textit{BSDAR} & \bf 0.405  & \bf 0.423 & \bf 0.359  & \bf 0.408 & \bf 0.335 & \bf 0.349 & \bf 0.277  & \bf 0.285 \\
\hline
\end{tabular}
}
\caption{\label{table:ABS2} Abstractive Performance on longer sequences ($n-$grams, $n=3 \ldots 5$). Scores are based on \texttt{\small{ROUGE-L average F1-score}}. \small{\textit{$^{\dagger)}$ size of test corpora after preprocessing.}}}
\end{table*}

\subsection{Abstractive performance}
For a neural generation task, it is also important to maintain the model ability to generate abstractive sequences. The results in table \ref{table:ABS1} show that \textit{\textbf{BSDAR}} is able to maintain a high performance on generating sequences not featured in source text, given subsets with absent keyphrase references in a variable sequence length ($n=1, \ldots, 5$). We further evaluate the abstractive performance of the decoding algorithms on longer sequences (table \ref{table:ABS2}). Since this task is generally challenging for any models, we use ROUGE-L average $F_1-$score metric to match the prediction and ground truth set. In addition to maintaining a high abstractive performance in the subset with longer target sequences, \textit{\textbf{BSDAR}} is also able to disclose the actual intrinsic performance of \textbf{CorrRNN-L}. As compared to low decoding performance of \textbf{CorrRNN-L} based on standard beam search (BS), the decoding performance of \textbf{CorrRNN-L} based on \textit{\textbf{BSDAR}} is higher. This indicates that the corresponding model has actually learnt to attend on relevant words in source text, but the decoding algorithm fails to include the words in the final generated sequences. Understanding the intrinsic performance of a complex Seq2Seq model exemplified by this empirical result is becoming essential, specifically to further address a challenging abstractive generation task for any neural sequence models.

\begin{table*}
\centering
\resizebox{\textwidth}{!}{
\begin{tabular}{ | l | ccc | ccc | ccc | ccc | }
\hline
\bf Model & \multicolumn{3}{|c|}{\textbf{Inspec-L (189)$^{\dagger}$}} & \multicolumn{3}{|c|}{\textbf{Krapivin-L (118)$^{\dagger}$}} & \multicolumn{3}{|c|}{\textbf{NUS-L (121)$^{\dagger}$}} & \multicolumn{3}{|c|}{\textbf{Semeval2010-N (234)$^{\dagger}$} } \\
& \bf $R@10$ & \bf $R@50$ & ROUGE-L & \bf $R@10$ & \bf $R@50$ & ROUGE-L  & \bf $R@10$ & \bf $R50$ & ROUGE-L  & \bf $R@10$ & \bf $R@50$  & ROUGE-L \\
 \hline
TfIdf & 0.082 & 0.252 & 0.635 & \bf 0.079 & 0.189 & 0.562 & \bf 0.044 & 0.110 & 0.478 & 0.039 & 0.157 & \bf 0.441\\
CorrRNN $+$ BS & 0. & 0. & 0.256 & 0. & 0. & 0.265 & 0. & 0. & 0.251 & 0. & 0. & 0.236\\
CorrRNN $+$ BS++ & 0.005 & 0.005 &  0.294 & 0.011 & 0.011 & 0.306 & 0.009 & 0.009 & 0.299 & 0.006 & 0.006 & 0.261\\
CorrRNN $+$ \bf \textit{BSDAR} & 0.007 & 0.130 & 0.643 & 0.009 & 0.123 & 0.560 & 0.009  & 0.070  &  0.4960 & 0.009 & 0.058  & 0.426 \\
CorrRNN-L $+$ BS & 0. & 0. & 0.138 & 0. & 0. & 0.151 & 0. & 0. & 0.152 & 0. & 0. & 0.134 \\
CorrRNN-L $+$ BS++ & 0. & 0.003 & 0.308 & 0.009 & 0.009 & 0.341 & 0.005 & 0.005 & 0.320 & 0.002 & 0.003 & 0.299 \\
CorrRNN-L $+$ \bf \textit{BSDAR} & \bf 0.102 & \bf 0.342 & \bf 0.664 & 0.057 & \bf 0.219 & \bf 0.572 & 0.035  & \bf 0.164 & \bf 0.4962 & \bf 0.049 & \bf 0.162 &  0.436 \\
\hline
\end{tabular}
}
\caption{\label{table:bs_extractiveness1} Comparison between models in subsets with longer keyphrases (n-grams, $n=3 \ldots 5$). \textit{\small $^{\dagger)}$ size of test corpora after preprocessing.}} 
\end{table*}

\subsection{On sequence length bias issue}

To measure whether the proposed method in this study (\textit{\textbf{BSDAR}}) can overcome the algorithm bias to shorter sequences, we compare the Seq2Seq decoding performance based on \textit{\textbf{BSDAR}} with \textbf{Tf-Idf} unsupervised keyword extractor in a subset with longer target sequences ($n-$grams, $n>3$). Intuitively, this subset introduces challenges for both neural network and non-neural network (Tf-Idf) approaches, since both models may suffer with sequence length bias issues, resulting the prediction with shorter sequence length. The result in table \ref{tab:diversity_compar} shows that the proposed solution can improve the decoding performance of Seq2Seq models, outperforming Tf-Idf in three data sets, as compared to Seq2Seq with standard beam (BS) and heuristic-based only beam decoding (BS++). 

\subsection{On diversity issue}

We also show that \textit{\textbf{BSDAR}} can overcome the diversity issue of beam decoding algorithm. Based on the result in table \ref{tab:diversity_compar}, \textit{\textbf{BSDAR}} can maintain high diversity score (section \ref{sec:seqlen}) close to the diversity measure of ground truth references. Furthermore, as compared to the standard beam (BS) and heuristic beam (BS++), \textit{\textbf{BSDAR}} shows a reasonably better decoding performance for recalling uni-gram ($R^{(1)}$), bi-gram ($R^{(2)}$), and $n-$gram references ($R^{(n)}, n \geq 3$).

\begin{table}[!ht]
    \centering
    \resizebox{.45\textwidth}{!}{
    \begin{tabular}{|l|c | c |c |c |}
    \hline
    \bf Keyphrase set &  \texttt{\small DIV}$^{(n)}$ & \bf $R^{(1)}$ & \bf $R^{(2)}$ & \bf $R^{(n)}$ \\
    \hline
    Ground truth & 0.942 & N/A & N/A & N/A\\
    BS  & 0.058 & 0.120 & 0.017 & 0.\\
    BS++  & 0.661  & 0.112 & 0.030 & 0.004 \\
    \textit{\textbf{BSDAR}}  & \bf 0.746 & \bf 0.301 & \bf 0.198 & \bf 0.223 \\
    \hline
    
    \end{tabular}
    }
    \caption{Diversity measure of decoding algorithms across data sets}
    \label{tab:diversity_compar}
\end{table}

\section{Conclusion}

We present a simple approach based on word-level and ngram-level reward function to overcome two beam search decoding issues in neural keyphrase generation. We show empirically that the proposed \textbf{BSDAR} not only performs well on improving the generation of longer and diverse sequences, but also maintains the Seq2Seq generative ability on producing abstractive keyphrases.



\bibliography{acl2019}

\begin{thebibliography}{18}
\expandafter\ifx\csname natexlab\endcsname\relax\def\natexlab#1{#1}\fi

\bibitem[{Bahdanau et~al.(2016)Bahdanau, Brakel, Xu, Goyal, Lowe, Pineau,
  Courville, and Bengio}]{Bahdanau2017Actor}
Dzmitry Bahdanau, Philemon Brakel, Kelvin Xu, Anirudh Goyal, Ryan Lowe, Joelle
  Pineau, Aaron~C. Courville, and Yoshua Bengio. 2016.
\newblock \href {http://arxiv.org/abs/1607.07086} {An actor-critic algorithm
  for sequence prediction}.
\newblock \emph{CoRR}, abs/1607.07086.

\bibitem[{Bahdanau et~al.(2014)Bahdanau, Cho, and Bengio}]{Bahdanau2014Neural}
Dzmitry Bahdanau, Kyunghyun Cho, and Yoshua Bengio. 2014.
\newblock \href {http://arxiv.org/abs/1409.0473} {Neural machine translation by
  jointly learning to align and translate}.
\newblock \emph{CoRR}, abs/1409.0473.

\bibitem[{Chen et~al.(2018)Chen, Zhang, Wu, Yan, and Li}]{Chen2018Keyphrase}
Jun Chen, Xiaoming Zhang, Yu~Wu, Zhao Yan, and Zhoujun Li. 2018.
\newblock \href {http://aclweb.org/anthology/D18-1439} {Keyphrase generation
  with correlation constraints}.
\newblock In \emph{Proceedings of the 2018 Conference on Empirical Methods in
  Natural Language Processing}, pages 4057--4066. Association for Computational
  Linguistics.

\bibitem[{Cho et~al.(2014)Cho, Van~Merri{\"e}nboer, Gulcehre, Bahdanau,
  Bougares, Schwenk, and Bengio}]{cho2014learning}
Kyunghyun Cho, Bart Van~Merri{\"e}nboer, Caglar Gulcehre, Dzmitry Bahdanau,
  Fethi Bougares, Holger Schwenk, and Yoshua Bengio. 2014.
\newblock Learning phrase representations using rnn encoder-decoder for
  statistical machine translation.
\newblock \emph{arXiv preprint arXiv:1406.1078}.

\bibitem[{Gu et~al.(2016)Gu, Lu, Li, and Li}]{Jiatao2016Copying}
Jiatao Gu, Zhengdong Lu, Hang Li, and Victor~O.K. Li. 2016.
\newblock \href {http://www.aclweb.org/anthology/P16-1154} {Incorporating
  copying mechanism in sequence-to-sequence learning}.
\newblock In \emph{Proceedings of the 54th Annual Meeting of the Association
  for Computational Linguistics (Volume 1: Long Papers)}, pages 1631--1640,
  Berlin, Germany. Association for Computational Linguistics.

\bibitem[{Hulth(2003)}]{Hulth:2003}
Anette Hulth. 2003.
\newblock Improved automatic keyword extraction given more linguistic
  knowledge.
\newblock In \emph{Proceedings of the 2003 conference on Empirical methods in
  natural language processing}, pages 216--223.

\bibitem[{Jean et~al.(2015)Jean, Cho, Memisevic, and Bengio}]{jean2015Using}
Sebastien Jean, Kyunghyun Cho, Roland Memisevic, and Yoshua Bengio. 2015.
\newblock \href {http://www.aclweb.org/anthology/P15-1001} {On using very large
  target vocabulary for neural machine translation}.
\newblock In \emph{Proceedings of the 53rd Annual Meeting of the Association
  for Computational Linguistics and the 7th International Joint Conference on
  Natural Language Processing (Volume 1: Long Papers)}, pages 1--10, Beijing,
  China. Association for Computational Linguistics.

\bibitem[{Kim et~al.(2010)Kim, Medelyan, Kan, and Baldwin}]{Kim:2010}
Su~Nam Kim, Olena Medelyan, Min-Yen Kan, and Timothy Baldwin. 2010.
\newblock \href {http://dl.acm.org/citation.cfm?id=1859664.1859668}
  {Semeval-2010 task 5: Automatic keyphrase extraction from scientific
  articles}.
\newblock In \emph{Proceedings of the 5th International Workshop on Semantic
  Evaluation}, SemEval '10, pages 21--26, Stroudsburg, PA, USA. Association for
  Computational Linguistics.

\bibitem[{Krapivin et~al.(2008)Krapivin, Autayeu, and Marchese}]{Krapivin:2009}
Mikalai Krapivin, Aliaksandr Autayeu, and Maurizio Marchese. 2008.
\newblock Large dataset for keyphrases extraction.
\newblock Technical Report DISI-09-055, DISI, Trento, Italy.

\bibitem[{Li and Jurafsky(2016)}]{LiJurafskiMutual2016}
Jiwei Li and Dan Jurafsky. 2016.
\newblock \href {http://arxiv.org/abs/1601.00372} {Mutual information and
  diverse decoding improve neural machine translation}.
\newblock \emph{CoRR}, abs/1601.00372.

\bibitem[{Meng et~al.(2017)Meng, Zhao, Han, He, Brusilovsky, and
  Chi}]{Meng2017Deep}
Rui Meng, Sanqiang Zhao, Shuguang Han, Daqing He, Peter Brusilovsky, and
  Yu~Chi. 2017.
\newblock \href {http://aclweb.org/anthology/P17-1054} {Deep keyphrase
  generation}.
\newblock In \emph{Proceedings of the 55th Annual Meeting of the Association
  for Computational Linguistics (Volume 1: Long Papers)}, pages 582--592,
  Vancouver, Canada. Association for Computational Linguistics.

\bibitem[{Nguyen and Kan(2007)}]{Nguyen:2007}
Thuy~Dung Nguyen and Min-Yen Kan. 2007.
\newblock Key phrase extraction in scientific publications.
\newblock In \emph{Proceeding of International Conference on Asian Digital
  Libraries}, pages 317--326.

\bibitem[{Ranzato et~al.(2016)Ranzato, Chopra, Auli, and
  Zaremba}]{Ranzato2016Sequence}
Marc'Aurelio Ranzato, Sumit Chopra, Michael Auli, and Wojciech Zaremba. 2016.
\newblock \href {http://arxiv.org/abs/1511.06732} {Sequence level training with
  recurrent neural networks}.
\newblock In \emph{4th International Conference on Learning Representations,
  {ICLR} 2016, San Juan, Puerto Rico, May 2-4, 2016, Conference Track
  Proceedings}.

\bibitem[{See et~al.(2017)See, Liu, and Manning}]{See2017Get}
Abigail See, Peter~J. Liu, and Christopher~D. Manning. 2017.
\newblock \href {https://doi.org/10.18653/v1/P17-1099} {Get to the point:
  Summarization with pointer-generator networks}.
\newblock In \emph{Proceedings of the 55th Annual Meeting of the Association
  for Computational Linguistics (Volume 1: Long Papers)}, pages 1073--1083,
  Vancouver, Canada. Association for Computational Linguistics.

\bibitem[{Sutskever et~al.(2014)Sutskever, Vinyals, and
  Le}]{sutskever2014sequence}
Ilya Sutskever, Oriol Vinyals, and Quoc~V Le. 2014.
\newblock Sequence to sequence learning with neural networks.
\newblock In \emph{Advances in neural information processing systems}, pages
  3104--3112.

\bibitem[{Vinyals et~al.(2015)Vinyals, Toshev, Bengio, and
  Erhan}]{vinyals2015show}
Oriol Vinyals, Alexander Toshev, Samy Bengio, and Dumitru Erhan. 2015.
\newblock Show and tell: A neural image caption generator.
\newblock In \emph{Proceedings of the IEEE conference on computer vision and
  pattern recognition}, pages 3156--3164.

\bibitem[{Wiseman and Rush(2016)}]{wiseman2016sequence}
Sam Wiseman and Alexander~M. Rush. 2016.
\newblock \href {https://doi.org/10.18653/v1/D16-1137} {Sequence-to-sequence
  learning as beam-search optimization}.
\newblock In \emph{Proceedings of the 2016 Conference on Empirical Methods in
  Natural Language Processing}, pages 1296--1306, Austin, Texas. Association
  for Computational Linguistics.

\bibitem[{Xu et~al.(2015)Xu, Ba, Kiros, Cho, Courville, Salakhudinov, Zemel,
  and Bengio}]{xu2015show}
Kelvin Xu, Jimmy Ba, Ryan Kiros, Kyunghyun Cho, Aaron Courville, Ruslan
  Salakhudinov, Rich Zemel, and Yoshua Bengio. 2015.
\newblock Show, attend and tell: Neural image caption generation with visual
  attention.
\newblock In \emph{International conference on machine learning}, pages
  2048--2057.

\end{thebibliography}
\bibliographystyle{acl_natbib}

\end{document}